\documentclass[acmlarge,screen,nonacm]{acmart}

\AtBeginDocument{%
  \providecommand\BibTeX{{%
    \normalfont B\kern-0.5em{\scshape i\kern-0.25em b}\kern-0.8em\TeX}}}

\usepackage{pdfrender}
\newcommand*{\boldcheckmark}{%
  \textpdfrender{
    TextRenderingMode=FillStroke,
    LineWidth=.5pt, 
  }{\checkmark}%
}

\begin{document}

\title{Video Ads Content Structuring by Combining \\ Scene Confidence Prediction and Tagging}
\titlenote{Technical report for the Tencent Advertising Algorithm Competition 2021 (awarded for the best overseas team solution)}

\author{Tomoyuki Suzuki}
\email{suzuki_tomoyuki@cyberagent.co.jp}
\affiliation{%
  \institution{CyberAgent, Inc.}
  \streetaddress{Shibuya Scramble Square 2-24-12 22F}
  \city{Shibuya}
  \state{Tokyo}
  \country{JAPAN}
  \postcode{150-6122}
}

\author{Antonio Tejero-de-Pablos}
\email{antonio_tejero@cyberagent.co.jp}
\affiliation{%
  \institution{CyberAgent, Inc.}
  \streetaddress{Shibuya Scramble Square 2-24-12 22F}
  \city{Shibuya}
  \state{Tokyo}
  \country{JAPAN}
  \postcode{150-6122}
  }

\renewcommand{\shortauthors}{Suzuki, et al.}

\begin{abstract}
  Video ads segmentation and tagging is a challenging task due to two main reasons: (1) the video scene structure is complex and (2) it includes multiple modalities (e.g., visual, audio, text.). While previous work focuses mostly on activity videos (e.g. ``cooking'', ``sports''), it is not clear how they can be leveraged to tackle the task of video ads content structuring. In this paper, we propose a two-stage method that first provides the boundaries of the scenes, and then combines a confidence score for each segmented scene and the tag classes predicted for that scene. We provide extensive experimental results on the network architectures and modalities used for the proposed method. Our combined method improves the previous baselines on the challenging ``Tencent Advertisement Video'' dataset.
  
\end{abstract}




\maketitle

\section{Introduction}
The field of video content structuring has applications in many areas related to multimedia, such as video recommendation, video highlight detection and smart surveillance.
This field is very challenging, in particular the task of video ads content structuring. Compared to general video understanding problems, ad videos include more complex interactions. They are like short movies in which actors are involved in a situation related to a product/service. Furthermore, ad videos contain multiple modalities (e.g., audio, video). In order to understand its content, a typical methodology is to segment the video in \textit{scenes}, and tag each scene with multiple labels that represent where, how, and what is going on (e.g., ``place'', ``style'', ``presentation''). In addition, a scene is composed of several subscenes that share the same overall semantics but differ in the camera angle and other specifics; these are called ``shots''.

\subsection{Related work}

Traditionally, general video segmentation methods rely on low level features (e.g., color histograms, HOG, etc.) to detect when the content changed. This strategy is valid for shots detection, but ads video segmentation requires higher semantics in their visual cues in order to distinguish when a scene ends and the next start. This is because, as the camera and other elements may abruptly change in the scene, low level visual features alone are unreliable.
In~\cite{rao2020local}, a method for modeling scene boundaries is proposed. By leveraging multiple semantics (i.e., place, cast, action, audio) their method can model the transitions between scenes. This is made in a local-to-global fashion, in which the different shots of the clip are first detected, and then each one is classified as a transition between scenes or not.
Following a similar strategy, the method in~\cite{tapu2020deep} first divides the video into shots using low-level visual features, and then clusters these shots into scenes by leveraging visual, audio and higher-semantics (i.e., object and person detection).
Given a dense set of proposals for the boundaries between scenes in a temporally segmented video, selecting the correct initial and final boundaries is not trivial. To tackle this, a method for assigning a confidence score to the proposals is presented in~\cite{lin2019bmn}.

These methods can provide state-of-the-art results in a variety of video temporal segmentation tasks, but their application to ad video has not been studied yet. Moreover, it is not clear which combination of features is the most effective when applying these methods.

In our work, we propose a method that combines \textit{temporal segmentation}, \textit{scene proposal scoring} and \textit{tagging}. We design several network combinations and evaluate their effectiveness on the challenging ``Tencent Advertisement Video'' dataset. Furthermore, we evaluate the use of several modalities (e.g., video, audio and text) and their representations (e.g., pretrained, learned).

\begin{figure}[tbp]
 \centering
 \includegraphics[keepaspectratio, scale=0.55]{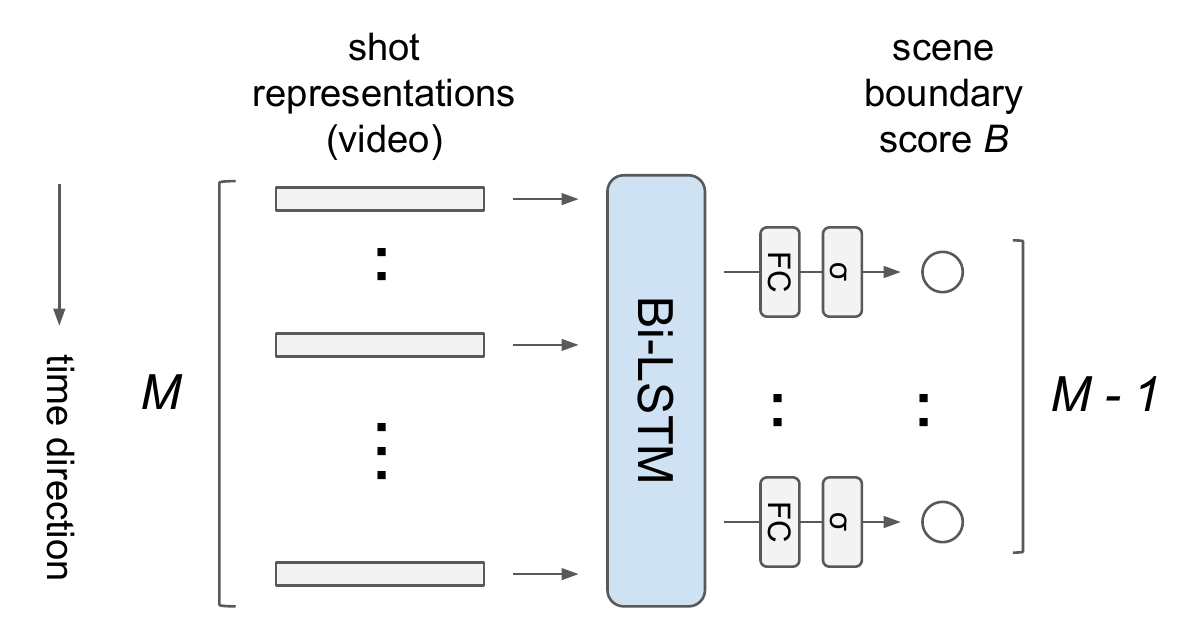}
 \caption{BoundaryNet. FC and $\sigma$ represent "fully-connected layer" and "sigmoid function" respectively.}
 \label{fig:boundary_net}
\end{figure}

\begin{figure}[tbp]
 \centering
 \includegraphics[keepaspectratio, scale=0.55]{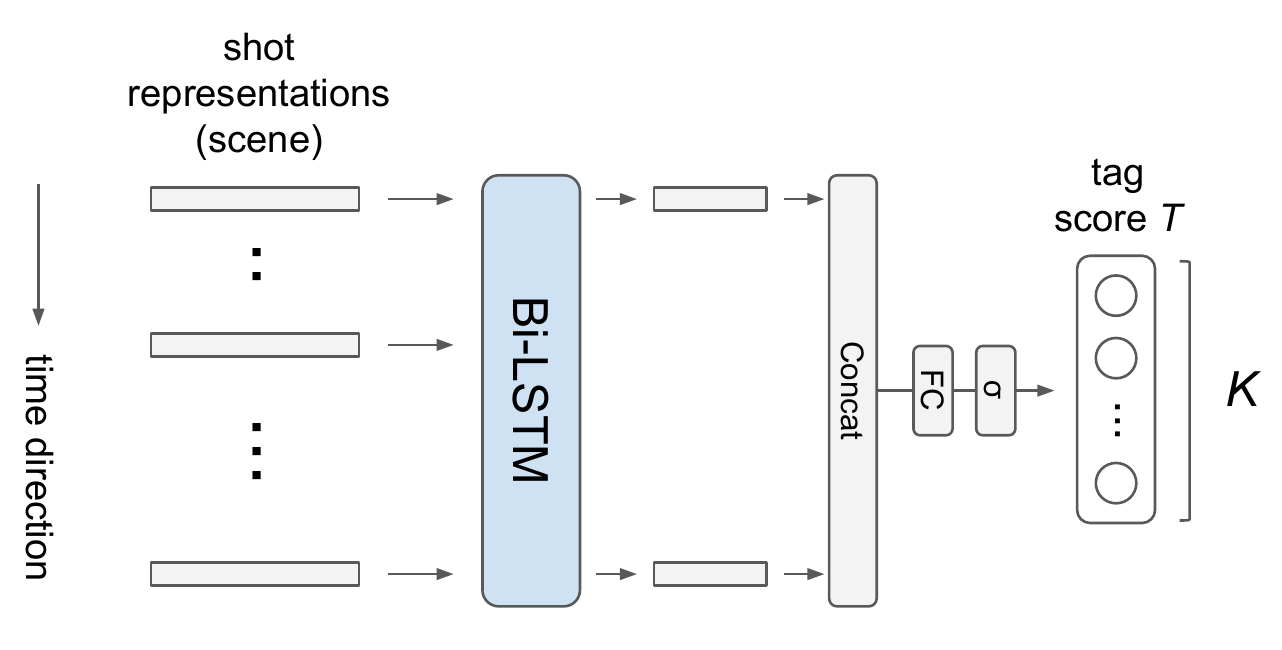}
 \caption{TagNet. FC and $\sigma$ represent "fully-connected layer" and "sigmoid function" respectively.}
 \label{fig:tag_net}
\end{figure}

\section{Method}

\subsection{Overview}


Following~\cite{rao2020local}, we adopt a shot-based model, which takes "shots" as the minimum temporal input unit, and outputs "scenes" (start and end time stamps) and scores of tags corresponding to each scene.
In this task, a "shot" is an unbroken sequence of frames recorded from the same camera, while a "scene" is a semantic temporal unit. Therefore, this approach assumes most shots can be uniquely categorized into one scene. We detect shots by the off-the-shelf method~\cite{sidiropoulos2011temporal}. In the following sections, we present our solution in detail.
\subsection{Shot representation}
\label{sec:representation}

Ad videos contain various multi-modal information. We extract the following representations from each shot and fuse them as a shot representation.
\ \\
\textbf{Pretrained visual representation.} We extract the hidden layer activations of pretrained CNNs for 3 keyframes (center, start and end) in each shot and average them to get a pretrained visual representation.
Specifically, we utilize ResNet-50~\cite{he2016deep} and Inception-V3~\cite{szegedy2016rethinking} pretrained on ImageNet~\cite{deng2009imagenet}. 
For both models, we extract activations from the layer before the last classification layer. Then, only for Inception-V3, we reduce their dimensionality via PCA (as in~\cite{abu2016youtube}). The PCA parameters are fitted on train split of YouTube8M~\cite{abu2016youtube}.
Finally, we obtain a 2048-D and a 1024-D activation vector respectively for each shot.
\ \\
\textbf{Image.}
In addition to the pretrained visual representation extracted via the above frozen CNNs, we also take the raw center-frame image in each shot and encode it via ResNet-18~\cite{he2016deep}, which is trained end-to-end.
\ \\
\textbf{Audio.}
Audio features are obtained by converting the audio signal in each shot into a mel-spectrogram using STFT~\cite{umesh1999fitting} with a sampling rate of 16K Hz and a windowed signal length of 512, and then encoding it through a 6-layer light-weight CNN. As the ResNet-18 above, this CNN is trained end-to-end.
\ \\
\textbf{Text.}
We also extract the text features from 3 key frames (center, start and end) in each shot. We extract the subtitles and other text in each keyframe via PaddleOCR~\cite{du2020pp}, and encode it via a pretrained BERT feature extractor for text in Chinese language~\cite{cui-etal-2020-revisiting}. This model outputs a feature vector per symbol in the input sentence(s), and the final representation is the average pooling of all vectors.
\ \\
\textbf{Shot length.}
Although a shot has a variable temporal duration, the above-mentioned representations omit it. Therefore, we represent it as a scalar value (in seconds).

Finally, we concatenate the above representations to obtain a shot representation.

\subsection{Modeling}
\label{sec:modeling}
\begin{figure}[tbp]
 \centering
 \includegraphics[keepaspectratio, scale=0.55]{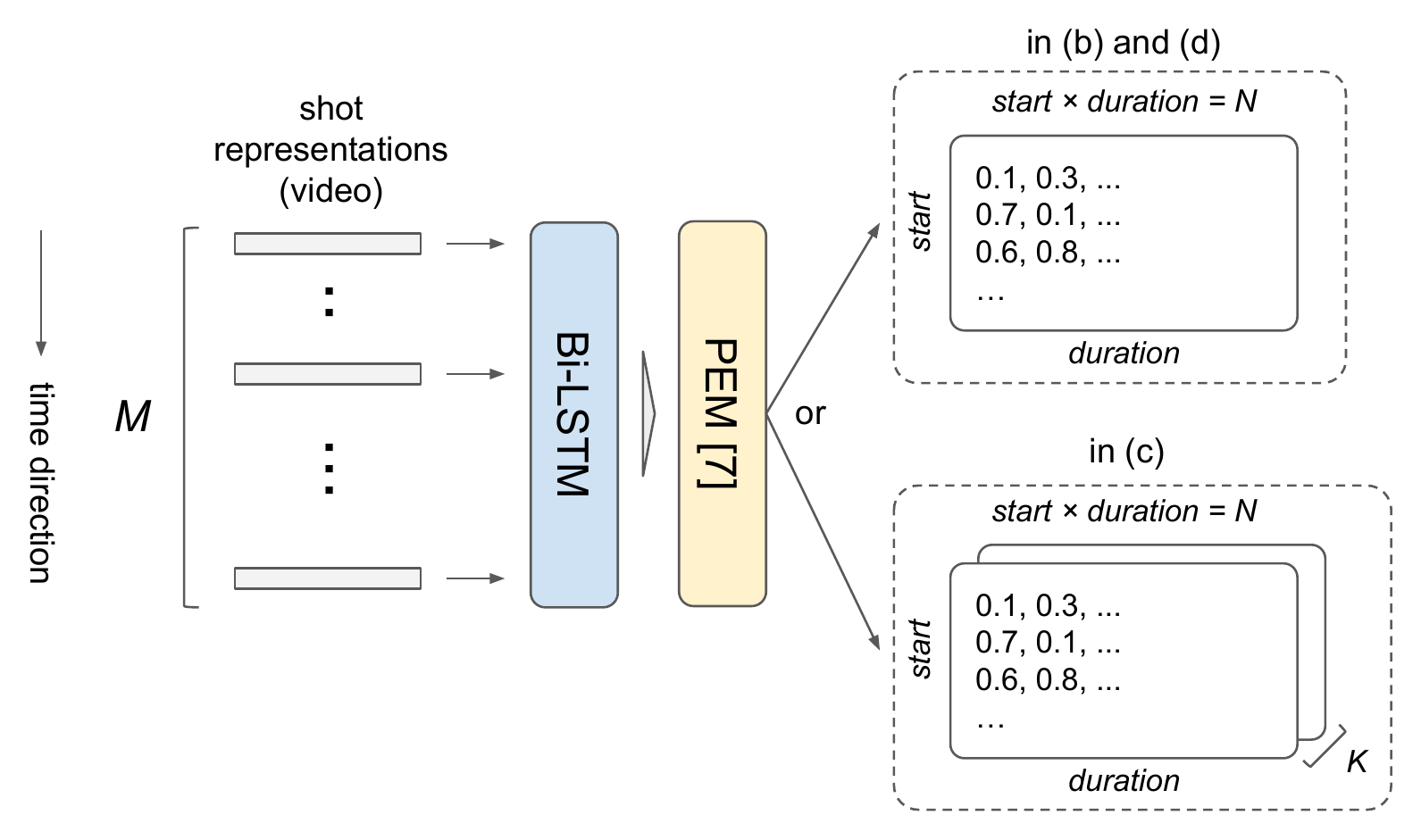}
 \caption{SegmentNet.}
 \label{fig:segment_net}
\end{figure}

In order to approach the task of Video Ads Content Structuring, we present four different models that are a combination of three submodel blocks.

\subsubsection{Submodel blocks}
\ \\
\textbf{BoundaryNet} (Figure~\ref{fig:boundary_net}) classifies whether each shot boundary (the boundary between two consecutive shots) in a video is a scene boundary (the boundary between two consecutive scenes) or not.
This submodel consists of a 2-layer bi-directional LSTM followed by fully-connected layer. Then, a sigmoid function outputs a scene boundary score in the range of $[0, 1]$ for each shot boundary as:
\begin{equation}
  B=\{b_m|m\in [1,M -1]\},
\end{equation}
where $M$ is the number of shots in a video, and therefore $M - 1$ is the number of shot boundaries.
It is trained by minimizing a cross-entropy loss. This submodel can be seen as a simple version of LGSS~\cite{rao2020local}, without BNet and Global Optimal Grouping.

\textbf{SegmentNet} (Figure~\ref{fig:segment_net}) is based on the proposal evaluation module (PEM) of BMN~\cite{lin2019bmn} and differs from the original only in the base module for considering the video global context. Specifically, we replaced the base module in ~\cite{lin2019bmn}) (i.e., a temporal convolution) with a 2-layer bi-directional LSTM.
From an input sequence of shot representations, this submodel predicts a score(s) for every scene in a video as below:
If SegmentNet only solves scene segmentation ((b) in Section~\ref{sec:overall_models}), the output for each scene is a scalar value in the range $[0,1]$:
\begin{equation}
  S=\{s_n|n\in [1,N]\},
\end{equation}
where N is the number of scenes predicted.
On the other hand, if it is employed for solving both scene segmentation and tagging ((c) in Section~\ref{sec:overall_models}), the output for each scene is a $T$-dimensional vector in the range of $[0,1]$:
\begin{equation}
  S=\{s_{nk}|n\in [1,N],k\in [1,K]\},
\end{equation}
where $K$ is the number of tags.

\textbf{TagNet} (Figure~\ref{fig:tag_net}) predicts the score of each tag (i.e., class label) corresponding to a given segmented scene.
This model consists of a two-layer bi-directional LSTM and a fully connected layer.
First, a sequence of shot representations representing a scene is input to the bi-directional LSTM, and by combining the features at both temporal ends (start and end) of the output series, a fixed-length output is obtained for a variable-length input.
This output is then passed through the fully connected layer and the sigmold function, resulting in a K-dimensional vector with scores in the range of $[0,1]$:
\begin{equation}
  T=\{t_k|k\in [1,K]\},
\end{equation}
where K is the number of classes or \textit{tags}. Each score $t_k$ represents the probability of the scene to contain a certain class.
This submodel is trained by minimizing the cross entropy loss.

We train all submodels independently.


\subsubsection{Overall model configurations}
\label{sec:overall_models}
\ \\
\textbf{(a) BoundaryNet + TagNet.}
First, the scene boundaries are detected via BoundaryNet by thresholding its output for each shot (i.e., the shot is a boundary if $b_i\geq threshold_{b}$).
Then, for each segmented scene, TagNet predicts the score of each tag.
\\
\textbf{(b) SegmentNet + TagNet.}
First, the given sequence of shots is segmented into scenes by applying Non-Maximum-Suppression (NMS) to the predictions by SegmentNet.
Then, as in (a), TagNet predicts the tag scores for each scene.
Since the output of SegmentNet can be seen as the confidence of each segmented scene,  this is thought to contribute to improving the predicted rankings of tagged segments.
Therefore, in order to calculate the final tag score, we multiply the SegmentNet score $s_i$ and the TagNet score $t_i$ for each predicted scene segment.
\\
\textbf{(c) SegmentNet (single stage).}
SegmentNet directly predicts both the segmented scenes and the tag scores for each scene. As in (b), NMS is applied to the prediction of SegmentNet.
\\
\textbf{(d) BoundaryNet + SegmentNet + TagNet.}
As in (a), BoundaryNet performs the initial scene segmentation. In addition, for each obtained scene, a scene score and the tag scores for each class are predicted by SegmentNet and TagNet respectively.
Finally, as in (b), we used the product of the scene score and tag score as the final score.
Since the output of SegmentNet represents the confidence of each scene segment as mentioned in (b), this setting is expected to rank the tagged segments more accurately than (a), which uses only the output of TagNet for ranking the scene segments.


\section{Evaluation}

\subsection{Dataset}

The ``Tencent Advertisement Video'' dataset contains 10000 multimodal videos divided into train and test splits (5000 videos each). For ourt experiments, we picked up randomly 1000 videos from the train split for validation, and use the rest (4000 videos) for training. Apart from the video frames, the advertisements (ads) contain audio and overlay text in Chinese language. This text includes the subtitles of what is being said in the ad (narration and actors), but also other text cues related to whatever service or product is being advertised. The average length of the videos is $42.74\pm 14.16$ seconds.
Each video in the train split is labeled using timestamps that indicate when a scene finishes and the next starts, and there is no overlapping between scenes. Then, each scene contains a series of multiple tags, which represent the classes each scene belongs to (e.g., \textit{office}, \textit{surprise}, \textit{teacher}\footnote{Translated by the authors, since tags are also in Chinese.}). For the competition purposes, the test split is not labeled.

We preprocess the videos by reducing their resolution to $1/4$ and subsampling their framerate to 12fps. This procedure was selected empirically, in order to reduce the computational cost of our method while avoiding an excessive information loss.

\subsection{Metrics}
\label{sec:metrics}
In order to evaluate our method, we used the metric indicated by the ``Tencent Advertisement Algorithm Competition'', namely, the product of the average mean Average Precision (mAP) and the Boundary f1-score (B-f1).

The mAP is the mean value of the APs for all tag classes, where a predicted segment is a true positive if the temporal intersection over union (tIOU) with a ground-truth segment is greater or equal to a given threshold.
Then, the average mAP is defined as the average of all mAP values computed with tIOU thresholds between $0.5$ and $0.95$ with a stride of $0.05$.
The B-f1 is the f1-score of the predicted boundaries, where a predicted boundary is a true positive if the absolute temporal difference with its respective ground-truth boundary is less than $0.5$ seconds.

In addition, to validate the performance of the scene segmentation independently, we use the Scene f1-score (S-f1), which is f1-score of the predicted scenes. We consider a predicted segment as a true positive if the tIoU with its ground-truth segment is greater than $0.75$.
Then, to validate the performance of the tagging task independently, we use basic mean Average Precision (mAP) calculated by averaging the APs for all classes.

\subsection{Implementation details}
During training, we use ADAM~\cite{kingma2015adam} for optimization, with a learning rate of $0.01$ and a batch size of $32$.
We add dropout (drop rate of $0.5$) right after the concatenation of representations to suppress over-fitting.

During prediction, for all experiments we set the threshold of BoundaryNet to $threshold_b = 0.65$ (value chosen empirically using the validation set).
Since the dataset does not contain overlapping scenes, we set the threshold of tIoU in the NMS of SegmentNet to $0$.

For shot segmentation, we use ~\cite{sidiropoulos2011temporal}. As a result we obtain 59912 shots for the train split and 64845 shots for the test split.

\subsection{Results}

\begin{table}[]
\caption{The best top 10 modality combinations for tagging. "vis. rep. R50" refers to the visual representations of ResNet-50 and "vis. rep. I3" refers to the visual representation of Inception-V3.}
\label{tab:representation_ablation_tag}
\Description{ablation study of input representation}
\begin{tabular}{cccccc}
\hline
\textbf{vis. rep. R50} & \textbf{vis. rep. I3} & \textbf{image}  & \textbf{audio}     & \textbf{text}      &  \textbf{\begin{tabular}[c]{@{}c@{}} mAP   (val) \end{tabular}} \\ \hline
\boldcheckmark    & \boldcheckmark      &           & \boldcheckmark &           & \textbf{0.3740} \\
\checkmark    & \checkmark      &           &           &           & 0.3710 \\
\checkmark    & \checkmark      & \checkmark & \checkmark &           & 0.3686 \\
\checkmark    & \checkmark      & \checkmark &           &           & 0.3651 \\
\checkmark    & \checkmark      & \checkmark &           & \checkmark & 0.3648 \\
\checkmark    & \checkmark      & \checkmark & \checkmark & \checkmark & 0.3637 \\
\checkmark    & \checkmark      &           &           & \checkmark & 0.3600 \\
\checkmark    & \checkmark      &           & \checkmark & \checkmark & 0.3590 \\
\checkmark    &                &           & \checkmark &           & 0.3544 \\
\checkmark    &                &           &           &           &  0.3542 \\
\hline
\end{tabular}
\end{table}

\begin{table}[]
\caption{Modality combinations for scene segmentation. "vis. rep. R50" refers to the visual representations of ResNet-50 and "vis. rep. I3" refers to the visual representations of Inception-V3.}
\label{tab:representation_ablation_seg}
\begin{tabular}{ccccc}
\hline
\textbf{vis. rep. R50} & \textbf{vis. rep. I3} & \textbf{image} & \textbf{audio} & \textbf{ mAP (val)} \\ \hline
\boldcheckmark               &                         & \boldcheckmark      &                & \textbf{0.8685}             \\
\checkmark               &                         &                & \checkmark      & 0.8681             \\
\checkmark               &                         &                &                & 0.8645             \\
\checkmark               & \checkmark               & \checkmark      &                & 0.8612             \\
\checkmark               & \checkmark               & \checkmark      & \checkmark      & 0.8606             \\
\checkmark               & \checkmark               &                &                & 0.8573             \\
\checkmark               & \checkmark               &                & \checkmark      & 0.8568             \\
\checkmark               &                         & \checkmark      & \checkmark      & 0.8550              \\ \hline
\end{tabular}
\end{table}

\begin{table}[]
\caption{Comparison results of different models for video ads structuring. The same shot representations (vis. rep. R50 + vis. rep. I3 + image + audio) are used in all settings.}
\label{tab:modeling_ablation}
\Description{ablation study of modeling}
\begin{tabular}{cccccc}
\hline
\textbf{Model} & 
\textbf{\begin{tabular}[c]{@{}c@{}}avg. mAP \\  (val) \end{tabular}} &  
\textbf{\begin{tabular}[c]{@{}c@{}}B-f1 \\ (val) \end{tabular}} &  
\textbf{\begin{tabular}[c]{@{}c@{}}S-f1 \\ (val)\end{tabular}} &  
\textbf{\begin{tabular}[c]{@{}c@{}}avg. mAP \\ $\times$ B-f1 \\ (val) \end{tabular}} &
\textbf{\begin{tabular}[c]{@{}c@{}}avg. mAP \\ $\times$  B-f1 \\ (test)\end{tabular}} \\
\hline
(a) & 0.1885 & 0.7364 & 0.7263 & 0.1388 & 0.1204     \\
(b) & 0.1891 & 0.6791 & 0.6953 & 0.1284 & --     \\
(c) & 0.1099 & 0.6396 & 0.6584 & 0.0703 & --     \\
\textbf{(d)} & \textbf{0.1935} & \textbf{0.7364} & \textbf{0.7263} & \textbf{0.1425} & \textbf{0.1236} \\
\hline
\end{tabular}
\end{table}

We first study the effectiveness of combining different types of feature modalities for shot representation (those in Sec.~\ref{sec:representation}). We use the metrics from Sec.~\ref{sec:metrics} on the validation set, since labels for the test set are not available and the number of experiments for the competition is limited.
Table~\ref{tab:representation_ablation_tag} shows the top 10 combinations for the tagging task. The best combination uses both types of pretrained visual representations and audio. Although using all features may seem the best option, it does not provide the best results. In particular, the extracted text features contribute the least in comparison to the other modalities. We believe that the reason is because averaging the features reduces the representativeness of the features, and also using only three frames may no contain enough cues for prediction. Also, the pretrained visual representations alone perform better than in combination with the learned image features. This is due to the difficulty of training both modalities at the same time, since they have different optimization points.

We also conducted the same ablation study for the segmentation task (Table~\ref{tab:representation_ablation_seg}). Given the results of Table~\ref{tab:representation_ablation_tag}, we excluded the text modality and fixed the pretrained visual representation (ResNet-50). The best results are obtained with a different feature combination than the tagging task. In scene segmentation, visual cues (pretrained visual representation from ResNet-50 and image) were the most useful for determining scene boundaries in our method. This is expected, as in the ``Tencent Advertisement Video'' dataset the actor voices and background music do not always vary between scenes.


In addition, Table~\ref{tab:representation_ablation_tag} compares the model configurations described in Section~\ref{sec:modeling} for the task of scene tagging. Network (d) obtains the best results for all metrics, included the final evaluation metric on the test data. As expected, combining the tagging scores with the confidence of the scene itself provides the most accurate estimation of the segmentation and tag labels.

\section{Discussion and conclusion}

In our methodology, we opted for a light-weight pipeline composed of computationally non-expensive models.
We leverage pretrained features for video shot representation, and models with state-of-the-art performance in a variety of temporal segmentation tasks.
When evaluating our method, the experiments show that combining two scene segmentation models and solving segmentation and tagging separately is the most effective pipeline. Also, choosing a different set of feature modalities for segmentation and tagging achieves the optimal results.

In spite of using state-of-the-art methods, the numerical results obtained for the proposed metrics are not very high compared to the results obtained when the models were originally proposed. One reason is because advertisement videos have different characteristics than the datasets used in most video understanding works, such as activity videos or cooking videos. In particular, excluding some straight-forward labels such as \textit{outdoors} or \textit{promotion\_page}, several classes require understanding the profession of the actors and their relationship, which is not obvious from the image or audio. This problem is accentuated by the serious imbalance of some tag classes (e.g., \textit{doctor}).
Regarding the scene segmentation task, visual and audio remain the same among different scenes, which makes detecting their boundaries particularly hard.
These challenges could be overcome by leveraging the ``script'' (i.e., dialogue text, etc.) from the ads.
In any case, the criteria followed for labeling the video segments is unclear.

In this paper, we present our solution for the ``Tencent Advertisement Algorithm Competition''. We combine two state-of-the-art temporal segmentation methods with a tagging network in order to include scene confidence to the scene segmentation and tagging results. Our pipeline is light-weighted, and has been evaluated in different configurations of submodels and features representations.
As our future work, we plan to improve the representation of the text modality, as visual and audio are insufficient to solve this task. Additionally, we plan to provide a qualitative evaluation of the results obtained with the proposed method.

\begin{acks}
The authors would like to thank Mayu Otani, Yuki Iwazaki, and Kota Yamaguchi for their valuable advice.
\end{acks}

\bibliographystyle{ACM-Reference-Format}
\bibliography{biblio}






\end{document}